\documentclass[10pt,conference,a4paper]{IEEEtran}

\usepackage{times}
\usepackage{epsfig}
\usepackage{graphicx}
\usepackage{amsmath}
\usepackage{amssymb}
\usepackage{multirow}
\usepackage{enumitem}
\usepackage{subcaption}
\usepackage{comment}
\usepackage{adjustbox}
\usepackage{xcolor}
\usepackage{hyperref}

\ifCLASSINFOpdf
\else
\fi
\newcommand{\argmin}{\arg\!\min}
\newcommand{\widthscale}{0.11}
\newcommand{\eg}{\textit{e.g.}} 
\newcommand{\etc}{\textit{etc.}} 
\newcommand{\etal}{\textit{et al.}} 

\begin{document}

\title{Boosting High-Level Vision with Joint Compression Artifacts Reduction and Super-Resolution}

\author{\IEEEauthorblockN{Xiaoyu Xiang}
\IEEEauthorblockA{
Purdue University\\
West Lafayette, IN 47907, USA\\
xiang43@purdue.edu}
\and
\IEEEauthorblockN{Qian Lin}
\IEEEauthorblockA{HP Labs\\
Palo Alto, CA 94304, USA\\
qian.lin@hp.com}
\and
\IEEEauthorblockN{Jan P. Allebach}
\IEEEauthorblockA{Purdue University\\
West Lafayette, IN 47907, USA\\
allebach@purdue.edu}}

\maketitle

\begin{abstract}
Due to the limits of bandwidth and storage space, digital images are usually down-scaled and compressed when transmitted over networks, resulting in loss of details and jarring artifacts that can lower the performance of high-level visual tasks. In this paper, we aim to generate an artifact-free high-resolution image from a low-resolution one compressed with an arbitrary quality factor by exploring joint compression artifacts reduction (CAR) and super-resolution (SR) tasks. First, we propose a context-aware joint CAR and SR neural network (CAJNN) that integrates both local and non-local features to solve CAR and SR in one-stage. Finally, a deep reconstruction network is adopted to predict high quality and high-resolution images. Evaluation on CAR and SR benchmark datasets shows that our CAJNN model outperforms previous methods and also takes $26.2\%$ shorter runtime. Based on this model, we explore addressing two critical challenges in high-level computer vision: optical character recognition of low-resolution texts, and extremely tiny face detection. We demonstrate that CAJNN can serve as an effective image preprocessing method and improve the accuracy for real-scene text recognition (from $85.30\%$ to $85.75\%$) and the average precision for tiny face detection (from $0.317$ to $0.611$).
\end{abstract}
\section{Introduction}
Image down-scaling and compression techniques are widely used to meet the limits of hardware storage and data capacity, which sometimes sacrifice the visual effects as well as bringing troubles to visual detection and recognition. Compression artifact reduction (CAR)~\cite{shen1998review} and single image super-resolution (SISR) \cite{allebach1996edge} have been used in manifold applications, \eg { }digital zoom on smartphones \cite{wronski2019handheld}, video streaming \cite{xiang2020zooming} and print quality enhancement \cite{xiao2017real, xiang2019blockwise} to restore a high-quality and high-resolution image. Since Dong \cite{dong2014learning} first proposed SRCNN that applied a three-layer convolutional neural network (CNN) for the SISR task, more and more works  \cite{lim2017enhanced, zhang2018residual, zhang2020residual} have explored how to make use of the deep neural networks (DNN) to achieve better image quality as measured by PSNR and SSIM \cite{wang2004image}, or better visual quality \cite{ledig2017photo, wang2018esrgan} as measured by other perceptual metrics \cite{johnson2016perceptual, ma2017learning}. 

\begin{figure}[tb]
\captionsetup[subfigure]{labelformat=empty}
\begin{center}
  \begin{subfigure}[b]{0.30\linewidth}
  \includegraphics[width=\linewidth]{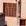}
  \subcaption{Input}
  \end{subfigure}
  \begin{subfigure}[b]{0.30\linewidth}
  \includegraphics[width=\linewidth]{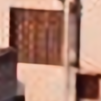}
  \subcaption{CAR+SR}
  \end{subfigure}
  \begin{subfigure}[b]{0.3\linewidth}
  \includegraphics[width=\linewidth]{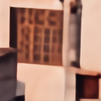}
  \subcaption{Ours (CAJNN)}
  \end{subfigure}
\end{center}
\vspace{-5mm}
\caption{Demonstration of the joint CAR and SR task. For a user's image without ground truth, our joint CAR and SR model (CAJNN) can generate better output with sharper edges and significantly fewer artifacts compared with the two-stage CAR+SR method.}
 \label{fig:teaser}
\vspace{-4mm}
\end{figure}

Conventional methods adopt a two-stage pipeline to leverage the quality and resolution of real-world images: first preprocess the user’s photos with a compression artifacts reduction (CAR) algorithm \cite{dong2015compression,galteri2017deep,zhang2008adaptive,zhang2018one,zhang2017beyond}, and then conduct a super-resolution (SR) algorithm \cite{allebach1996edge, dong2014learning, lim2017enhanced, zhang2018residual, zhang2020residual,zhang2018image, atkins2001optimal, dong2016accelerating,shi2016real}. However, the CAR step often causes loss of high-frequency information, which results in a lack of detail in reconstructed SR images. Besides, the computation and data transmission between the two models is time-consuming. To deal with these issues, a single-stage method that jointly solves the Compression Artifact Reduction and Super-Resolution (CARSR) problems is needed to reach a balance between reducing artifacts while retaining most details for the upscale step with a short run-time. 


Both CAR and SR aim to learn the high-frequency information for reconstruction. Thus, instead of simply concatenating two networks together, we design two functional modules in a single-stage network that reduces the model size by simplifying the two reconstruction processes into one, and can directly obtain high-quality SR output without reconstructing the intermediate artifact-free LR images. Towards this end, we propose a context-aware joint CAR and SR neural network (CAJNN) that can make use of the locally related features in low-quality, low-resolution images to reconstruct high-quality, high-resolution images. To train this network, we construct a paired LR-HR training dataset based on modeling the degradation kernels of web images. Our model turns out to be able to reconstruct high-resolution and artifact-free images with high stability for user’s images from a garden variety of web-apps (\eg { }Facebook, Instagram, WeChat). Figure \ref{fig:teaser} illustrates the performance of our proposed algorithm and the benefits of the single-stage joint CARSR method compared with previous two-stage methods: our result can reconstruct a more visually appealing output with accurate structures, sharp edges, and significantly fewer compression artifacts. These output images are not only more recognizable for human viewers, but also for off-the-shelf computer vision algorithms. In this paper, we demonstrate that our proposed CAJNN can enhance the detection and recognition accuracy of high-level vision tasks by reducing the compression artifacts and increasing the resolution of input images.

To summarize, our contributions are mainly three-fold:
(1) We propose a novel CAJNN framework that jointly solves the CAR and SR problems for real-world images, that are from unknown devices with unknown quality factors. Here, we explore ways to represent and combine both local and non-local information to enforce image reconstruction performance without knowing the input quality factor.
(2) Our experiments show that CAJNN achieves the new state-of-the-art performance on multiple datasets \eg { } Set5 \cite{bevilacqua2012low}, Set14 \cite{zeyde2010single}, BSD100 \cite{martin2001database}, Urban100 \cite{huang2015single}, \etc { }as measured by the PSNR and SSIM \cite{wang2004image} metrics. Compared with the prior art, it generates more stable and reliable outputs for any level of compression quality factors.
(3) We provide a new idea for enhancing high-level computer vision tasks like real-scene text recognition and extremely tiny face detection: by preprocessing the input data with our pretrained model, we can improve the performance of existing detectors. Our model demonstrates its effectiveness on the WIDER face dataset \cite{yang2016wider} and the ICDAR2013 Focused Scene Text dataset \cite{karatzas2013icdar}.

\section{Related Work}
\begin{description}[style=unboxed,leftmargin=0cm]
\item[CNN-based Single Image Super-Resolution]
Convolutional Neural Network (CNN) methods have demonstrated a remarkable capability to recover LR images with known kernels after the pioneering work of Dong \etal { }\cite{dong2014learning} that adopted a 3-layer CNN to learn an end-to-end mapping from LR images to HR images. The follow-up work FSRCNN \cite{dong2016accelerating} established the general structure of most SR networks until today, which conducts most computations in the low-resolution domain and upsamples the image to the required scale at the end of the network. After 2016, more and more works began to explore how to make the network go deeper. EDSR \cite{lim2017enhanced} reduces the number of parameters by removing the batch normalization layer, and shares the parameters between the low-scale and high-scale models to achieve better training results. RDN \cite{zhang2018residual, zhang2020residual} and RRDB \cite{wang2018esrgan} employ densely-connected residual groups as the major reconstruction block to reach large depth and to allow sufficient low-frequency information to be bypassed. In the meantime, some useful structures have been introduced to enhance the processing speed or output quality. Shi \etal { } \cite{shi2016real} designed a sub-pixel upscaling mechanism. RCAN \cite{zhang2018image} introduces a channel attention mechanism to rescale channel-wise features adaptively, and SAN \cite{dai2019second} exploits a more powerful feature expression with second-order channel attention.

\item[Compression Artifacts Reduction] Lossy compression methods~\cite{doulamis1998low, wu2017digital} are widely applied in web image transmission due to their higher compression rates. Traditional methods for the CAR problem generally fall into two categories: unsupervised methods, which include removing noise and increasing sharpness \cite{zhang2008adaptive}, and supervised methods like dictionary-based algorithms \cite{liu2015data}. After the success of SRCNN on the super-resolution task, Yu \etal{ }\cite{yu2016deep} directly applied its architecture to compression artifacts suppression. Similar to the development of SR, CNN-based CAR networks also go deeper with the introduction of residual blocks and skip connections \cite{chen2018cisrdcnn,svoboda2016compression,zini2019deep}. Besides, SSIM loss is employed \cite{galteri2017deep} as a supervision method to obtain better performance than MSE loss. JPEG-related priors are also considered in the network structure design, \eg { }DDCN \cite{guo2016building} adds a Discrete Cosine Transform (DCT)-domain before the dual networks, and the D$^3$ \cite{wang2016d3} takes a further step in the practice of dual-domain approaches \cite{liu2015data} by converting sparse-encoding approaches into a one-step sparse inference module.
\end{description}

Unlike the above approaches that require reconstructing the intermediate clean LR images, our joint CARSR framework directly obtains artifact-free HR images without prior information of quality factors or explicit CAR supervision in the LR domain.

\section{Joint Compression Artifacts Reduction and Super-Resolution}
Given an LR JPEG image $I^{LRLQ}$, our goal is to reconstruct the high resolution, high-quality image $G(I^{LRLQ})$ that approaches the high-resolution, high-quality ground truth $I^{HRHQ}$ with a generator $G$. The CARSR task can be expressed as:
\begin{equation}\label{eq:target}
    \argmin_{\theta}l(I^{HRHQ}, G(I^{LRLQ}, \theta_g)),
\end{equation}
where $l$ is any designated loss function (\eg { }MSE, L1, Charbonnier, \etc ). $G$ is the function representing our deep neural network with parameters $\theta$, for which we wish $G_{\theta} \approx F^{-1}((I^{LRLQ}\otimes k)\uparrow_s, q)$, where $\otimes$ stands for the convolution operation, $k$ is the degradation kernel of downsampling method, and $s$ is the downscaling factor. To effectively handle the CARSR task, we propose a single-stage framework, CAJNN. Our proposed model is end-to-end trainable with $I^{HRHQ}$ and $I^{LRLQ}$ pairs according to the function above.

\begin{figure}
    \centering
    \includegraphics[width=1.0\columnwidth]{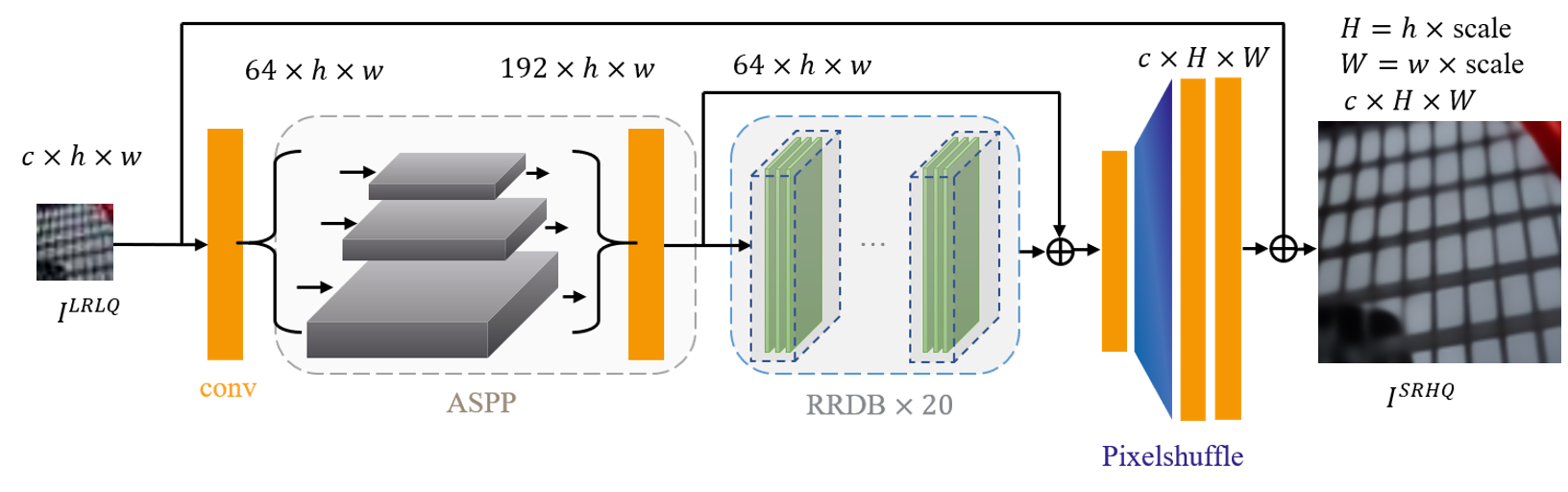}
    \caption{The network architecture of our proposed CAJNN. It directly reconstructs artifact-free HR images from the LR low-quality images $I^{LRLQ}$. Atrous Spatial Pyramid Pooling (ASPP) is adopted to utilize the inter-block features and intra-block contexts for the joint CARSR task. The reconstruction module turns the features into a deep feature map, which is converted to a high-quality SR output $I^{SRHQ}$ by the upsampling and enhancement module.}
    \label{fig:network}
\end{figure}

The CAJNN framework mainly consists of three modules (see Figure \ref{fig:network}): the \textit{context-aware feature extractor}, the \textit{reconstruction module}, and the \textit{upsampling and enhancement module}. The context-aware feature extractor captures and assembles both intra- and inter-block information from different receptive fields. The reconstruction module further refines the extracted feature maps. Finally, after the processing of the upsampling and enhancement module, these feature maps are converted to high-resolution outputs.


\subsection{Model}
Here we discuss the CAJNN structure in detail. The majority of our network operates in the feature domain. Given $I^{LRLQ}$ ($c \cdot h \cdot w$ in size), a feature extraction layer first turns the image into feature maps ($n_f \cdot h \cdot w$ in size, where $n_f$ denotes the number of feature channels) in the domain for the following process. The feature map will be converted to a high-resolution image ($c \cdot H \cdot W$ in size) after passing the upsampling and enhancement module. To achieve a balance between GPU capacity and output quality, we apply $n_f = 64$ channels to ensure enough information for the reconstruction. We adopt a $3\times3$ convolution layer that serves as the initial feature extractor. After this module, the input image $I^{LRLQ}$ is turned into a $64\cdot h\cdot w$ tensor $f^{L}$.

\subsubsection{Context-aware Feature Extractor}



The pipeline of JPEG compression involves the following steps: color space transformation (\eg { }JPEG, H.264/AVC, H.265/HEVC), downsampling, block splitting, discrete cosine transform (DCT), quantization, and entropy encoding. Some previous research assumed that the quality factors of input images are known, and the original images are well-aligned by $8\times 8$ patches with respect to the JPEG block boundaries. However, real-world inputs do not always meet such assumptions. In the worst case, the input images might be compressed multiple times and contain sub-blocks or larger blocks, which requires the model to be insensitive, or even blind to the encoding block alignment. Thus, the spatial context information of both intra- and inter- JPEG blocks is essential for designing a CARSR network.


We adopt the ASPP module to extract and integrate multi-scale features with an atrous spatial pooling pyramid (ASPP) \cite{chen2017deeplab}. We adjust the dilation rates of each layer in the pyramid to extend the filter's perceptive field for extracting different ranges of context information, in which the largest field-of-view should cover the $8 \times 8$ block. Besides, we should avoid sampling overlap in different levels of the $3\times 3$ convolutions. Considering the factors above, we choose 1, 3, 4 as the dilation groups to find a good balance between accurately retrieving local details and assimilating context information between adjacent blocks. The input tensors are sent to 3 layers of the pyramid: a $3\times 3$ convolutional layer with dilation rate = 1, a $3\times 3$ convolutional layer with dilation rate = 3, and a $3\times 3$ convolutional layer with dilation rate = 4. The outputs of these three layers are concatenated and aggregated by a $1\times 1$ convolution. The process in ASPP can be described by:
\begin{equation}
    f^{L'} = [C_{3\times 3,1}\otimes f^L | C_{3\times 3,3}\otimes f^L | C_{3\times 3,4}\otimes f^L]\otimes C_{1\times 1, 1}, 
\end{equation}
where $f^{L'}$ denotes the output feature ($64\cdot h \cdot w$ in size), $C_{a\times a, r}$ represents the parameters of $a\times a$ convolution with dilation rate $r$, and $|$ is a concatenation operation.

\subsubsection{Reconstruction}

RRDB (residual-in-residual dense block) \cite{wang2018esrgan} is applied as the basic block for the reconstruction trunk. Compared with residual blocks, it densely connects the convolution layers to local groups while removing the batch normalization layer. In our network, the reconstruction module includes 20 RRDBs.

\subsubsection{Upsampling and Enhancement}

After the reconstruction module, the image feature is preprocessed by a $3\times 3$ convolution layer before the PixelShuffle layer \cite{shi2016real} for upsampling. The Pixelshuffle layer produces an HR image from LR feature maps directly with one upscaling filter for each feature map. Compared with the upconvolution, the PixelShuffle layer is $\log_2 s^2$ times faster in theory because of applying sub-pixel activation to convert most of the computations from the HR to the LR domain. The feature $f^{L''}$ is turned into a $c \cdot sh \cdot sw$ HR image by the PixelShuffle layer, which can be described by:
\begin{equation}
    I^{SR'} = PS(W_L\otimes f^{L''} + b_L),
\end{equation}
where $W_L$ denotes the convolution weights and $b_L$ the bias in the LR domain, $PS$ is a periodic shuffling operator for re-arranging the input LR feature tensor $f^{L''}$ ($c\cdot s^2\cdot h \cdot w$) to a HR tensor of shape $c \cdot sh \cdot sw$:
\begin{equation}
    PS(T)_{x,y,c} = T_{\lfloor x/s \rfloor, \lfloor y/s \rfloor, c \cdot s \cdot \text{mod}(y,s) + c \cdot s \cdot \text{mod}(x,s)}.
\end{equation}

Instead of directly outputting the high-resolution image, we process it through two $3\times 3$ convolution layers for further enhancement, and get $I^{SR} = C_{3\times 3, 1} \otimes(C_{3\times 3, 1} \otimes I^{SR'})$.

To make the major network focus on learning the high-frequency information in the input image, we bilinearly upsample the input LR image $I^{LRLQ}$ and add it to form the final output $G(I^{LRLQ}, \theta_g)$:
\begin{equation}
    G(I^{LRLQ}, \theta_g) = I^{LRLQ}\uparrow_s + I^{SR}.
\end{equation}

This long-range skip connection changes the target of our major network from directly reconstructing a high-resolution image to reconstructing its residual. By letting the low-frequency information of the input bypass the major network, it lowers the difficulty of reconstruction and increases the convergence speed of the network.

\section{Experiments and Analysis}
\subsection{Experimental Setup}
\begin{description}[style=unboxed,leftmargin=0cm]
\item[Training Dataset] In this paper, we choose the DIV2K dataset \cite{Agustsson_2017_CVPR_Workshops} (800 RGB images of 2k resolution) and Flick2K dataset \cite{timofte2017ntire} (2650 RGB images of 2k resolution) as our training set. To get the training pairs that approach the degradation kernels of web images, we first model the downsampling and compression types of popular web platforms. We also discover that adding severely compressed samples to the training set can improve the output quality in terms of PSNR, even for input images compressed with high-quality factors. Based on these pre-experimental results, the images of the training set are downsampled with a scaling factor $s=4$ and compressed by MATLAB \cite{MATLAB:2018b} with random quality factors from 10 to 100. Besides, we perform data augmentation on these images by randomly cropping, randomly rotating by $90^{\circ}$, $180^{\circ}$, and $270^{\circ}$, and randomly horizontal-flipping. As a result, each cropped image patch can have eight augmentation types at maximum. 

\item[Test Datasets] We compare the performance of our model and previous methods on Set5 \cite{bevilacqua2012low}, Set14 \cite{zeyde2010single}, BSD100 \cite{martin2001database}, Urban100 \cite{huang2015single} and Manga109 \cite{matsui2017sketch}. Each image is downscaled $\times 4$ and compressed with quality factors of 10, 20 and 40 to be consistent with previous works. 

\item[Implementation Details] Our network is trained on one Nvidia Titan Xp graphics card. The batch size is 36, and the patch size is 128 for ground truth and 32 for low-resolution input. We use Adam \cite{kingma2014adam} as the optimizer with a cosine annealing learning rate, in which the initial learning rate is $2e-4$, and the minimum learning rate is $1e-7$. The scheduler restarts every $2.5e5$ iterations. The network is trained for $1e6$ iterations in total.
\end{description}

\subsection{Results for Image Quality Assessment}

\begin{description}[style=unboxed,leftmargin=0cm] 
\item[Comparison with SOTA on Standard Test Sets] We compare the performance of CAJNN to the previous state-of-the-art (SOTA) methods on the standard test sets as mentioned above. We report the PSNR and SSIM \cite{wang2004image} of the Y channels in the test sets to be consistent with previous works. We also show the number of parameters and the inference time on Set5 in Table \ref{tab:sr_result}. Depending on the workflow for solving the CAR and SR problem, these methods can be categorized into the following three types: (1) \textit{SR:} directly use pretrained SR models. (2) \textit{CAR+SR:} the aforementioned two-stage method, which first removes the compression artifacts and then sends the output images to the SR model. (3) \textit{Joint CAR \& SR:} the single-stage method that jointly handles CAR and SR with one model. We report both the direct output and the self-ensembled \cite{zhang2018residual} output of our network.

According to Table \ref{tab:sr_result}, CAJNN significantly outperforms the existing methods for all QFs, and yielding the highest overall PSNR across all five datasets with Set5. The improvement is consistently observed on SSIM, as well. Moreover, our model is more light-weight than most of the current models, including one-stage and two-stage in summation, which results in faster inference speed on the same hardware (all the tests are conducted on one Nvidia Titan Xp graphics card). 

\begin{table*}
\caption{Quantitative comparison of applying SOTA SR methods, two-stage SR and CAR methods, and our CAJNN. The best two results are highlighted in \textcolor{red}{red} and \textcolor{blue}{blue} colors, respectively. Our method greatly outperforms all two-stage methods in terms of PSNR and SSIM, while having a relatively small model size and shorter runtime. The runtime (inference only) is measured on the entire Set5.} 
\resizebox{\textwidth}{!}{
\begin{tabular}{c|llcccccccccccc}
\hline
\multirow{2}{*}{QF} & \multirow{2}{*}{Method}        & \multirow{2}{*}{Network} & \multirow{2}{*}{Runtime (s)} & \multirow{2}{*}{Parameters (Million)} & \multicolumn{2}{c}{Set5} & \multicolumn{2}{c}{Set14} & \multicolumn{2}{c}{BSD100} & \multicolumn{2}{c}{Urban100} & \multicolumn{2}{c}{Manga109} \\
                    &                                &                          &                              &                                       & PSNR       & SSIM        & PSNR        & SSIM        & PSNR        & SSIM         & PSNR         & SSIM          & PSNR         & SSIM          \\ \hline
                    
\multirow{10}{*}{10} & \multirow{4}{*}{SR}           
 & Bicubic                     & -                         & -                                  & 23.99      & 0.6329      & 22.94       & 0.5513      & 23.33       & 0.5303       & 20.95        & 0.5182        & 21.94        & 0.6383        \\
 & & EDSR                     & 1.94                         & 43.1                                  & 23.41      & 0.6019      & 22.48       & 0.5272      & 22.96       & 0.5098       & 20.57        & 0.5006        & 21.53        & 0.6151        \\
                    &                                & RCAN                     & 2.04                         & \textcolor{blue}{16}                                    & 23.14      & 0.5733      & 22.29       & 0.5064      & 22.78       & 0.4984       & 20.36        & 0.4819        & 21.21        & 0.5878        \\
                    &                                & RRDB                     &\textcolor{blue}{ 0.65 }                        & 16.7                                  & 22.43      & 0.5223      & 22.86       & 0.5051      & 20.43       & 0.4940       & 20.43        & 0.4940        & 21.34        & 0.6075        \\ \cline{2-15}
                    & \multirow{2}{*}{CAR+SR}        & ARCNN+RRDB               & 3.20+0.65                   & 0.56+16.7                             & 24.21      & 0.6699      & 23.38       & 0.5774      & 23.63       & 0.5474       & 21.28        & 0.5466        & 22.36        & 0.6856        \\
                    &                                & DnCNN+RRDB               & 0.38+0.65                    & 0.06+16.7                             & 24.07      & 0.6434      & 23.13       & 0.5582      & 23.37       & 0.5324       & 21.04        & 0.5305        & 22.10        & 0.6532        \\ \cline{2-15}
                    & \multirow{2}{*}{Joint CAR\&SR}                                & CAJNN (ours)                     & \textcolor{red}{0.48}                         & \textcolor{red}{ 14.8 }                                  & \textcolor{blue}{25.04}      & \textcolor{blue}{0.7169}      & \textcolor{blue}{23.95}       & \textcolor{blue}{0.6028}      & \textcolor{blue}{23.84}       & \textcolor{blue}{0.5598}       & \textcolor{blue}{21.97}        & \textcolor{blue}{0.5977}        & \textcolor{blue}{23.29}        & \textcolor{blue}{0.7333}        \\
                    &                                & CAJNN (ours, self-ensembled)                    & 2.50                         & \textcolor{red}{ 14.8 }                                  & \textcolor{red}{25.14}      & \textcolor{red}{0.7202}      & \textcolor{red}{24.03}       & \textcolor{red}{0.6052}      & \textcolor{red}{23.88}       & \textcolor{red}{0.5610}       & \textcolor{red}{22.18}        & \textcolor{red}{0.6051}        & \textcolor{red}{23.44}        & \textcolor{red}{0.7377}        \\ \hline
\multirow{10}{*}{20} & \multirow{4}{*}{SR}  & Bicubic                     & -                         & -                                  & 25.32      & 0.6761      & 23.85       & 0.5870      & 24.14       & 0.5611       & 21.66        & 0.5526        & 22.84        & 0.6724        \\
 & &  EDSR                     & 1.94                         & 43.1                                  & 24.76      & 0.6490      & 23.59       & 0.5707      & 23.88       & 0.5482       & 21.38        & 0.5427        & 22.58        & 0.6549        \\
                    &                                & RCAN                     & 2.04                         & \textcolor{blue}{16}                                    & 24.44      & 0.6226      & 23.40       & 0.5502      & 23.65       & 0.5351       & 21.12        & 0.5234        & 22.14        & 0.6253        \\
                    &                                & RRDB                     &\textcolor{blue}{ 0.65 }                        & 16.7                                  & 24.65      & 0.6450      & 23.57       & 0.5661      & 23.79       & 0.5442       & 21.25        & 0.5365        & 22.38        & 0.6474        \\ \cline{2-15}
                    & \multirow{2}{*}{CAR+SR}        & ARCNN+RRDB               & 3.20+0.65                   & 0.56+16.7                             & 25.40      & 0.7082      & 24.30       & 0.6091      & 24.39       & 0.5755       & 22.02        & 0.5811        & 23.52        & 0.7172        \\
                    &                                & DnCNN+RRDB               & 0.38+0.65                    & 0.06+16.7                             & 25.55      & 0.6946      & 24.24       & 0.6001      & 24.28       & 0.5679       & 21.90        & 0.5732        & 23.24        & 0.6961        \\ \cline{2-15}
                    & \multirow{2}{*}{Joint CAR\&SR}                                & CAJNN (ours)                     & \textcolor{red}{0.48}                         & \textcolor{red}{ 14.8 }                                  & \textcolor{blue}{26.59}      & \textcolor{blue}{0.7604}      & \textcolor{blue}{25.03}       & \textcolor{blue}{0.6391}      & \textcolor{blue}{24.70}       & \textcolor{blue}{0.5924}       & \textcolor{blue}{23.06}        & \textcolor{blue}{0.6482}        & \textcolor{blue}{24.81}        & \textcolor{blue}{0.7783}        \\
                    &                                & CAJNN (ours, self-ensembled)                    & 2.50                         & \textcolor{red}{ 14.8 }                                  & \textcolor{red}{26.65}      & \textcolor{red}{0.7633}      & \textcolor{red}{25.10}       & \textcolor{red}{0.6404}      & \textcolor{red}{24.74}       & \textcolor{red}{0.5936}       & \textcolor{red}{23.28}        & \textcolor{red}{0.6550}        & \textcolor{red}{24.98}        & \textcolor{red}{0.7820}        \\ \hline
\multirow{10}{*}{40} & \multirow{4}{*}{SR}             & Bicubic                     & -                         & -                                  & 26.38      & 0.7154      & 24.55       & 0.6201      & 24.77       & 0.5898       & 22.26        & 0.5877        & 23.66        & 0.7081        \\
 & &  EDSR                     & 1.94                         & 43.1                                  & 26.01      & 0.6972      & 24.48       & 0.6120      & 24.62       & 0.5836       & 22.18        & 0.5893        & 23.73        & 0.7003        \\
                    &                                & RCAN                     & 2.04                         & \textcolor{blue}{16}                                    & 25.70      & 0.6726      & 24.30       & 0.5936      & 24.36       & 0.5704       & 21.86        & 0.5690        & 23.13        & 0.6673        \\
                    &                                & RRDB                     &\textcolor{blue}{ 0.65 }                        & 16.7                                  & 25.99      & 0.6958      & 24.50       & 0.6079      & 24.54       & 0.5804       & 22.10        & 0.5851        & 23.50        & 0.6918        \\  \cline{2-15}
                    & \multirow{2}{*}{CAR+SR}        & ARCNN+RRDB               & 3.20+0.65                   & 0.56+16.7                             & 26.65      & 0.7495      & 25.16       & 0.6424      & 25.06       & 0.6053       & 22.82        & 0.6235        & 24.68        & 0.7578        \\ 
                    &                                & DnCNN+RRDB               & 0.38+0.65                    & 0.06+16.7                             & 26.87      & 0.7403      & 25.15       & 0.6373      & 25.00       & 0.5995       & 22.78        & 0.6194        & 24.42        & 0.7404        \\ \cline{2-15}
                    & \multirow{2}{*}{Joint CAR\&SR}                               & CAJNN (ours)                     & \textcolor{red}{0.48} & \textcolor{red}{ 14.8 }                                  & \textcolor{blue}{28.05}      & \textcolor{blue}{0.7981}      & \textcolor{blue}{25.96}       & \textcolor{blue}{0.6729}      & \textcolor{blue}{25.43}       & \textcolor{blue}{0.6240}       & \textcolor{blue}{24.09}        & \textcolor{blue}{0.6962}        & 26.25        & \textcolor{blue}{0.8177}        \\
                    &                                & CAJNN (ours, self-ensembled)                    & 2.50                         & \textcolor{red}{ 14.8 }                                  & \textcolor{red}{28.16}      & \textcolor{red}{0.7993}      & \textcolor{red}{26.03}       & \textcolor{red}{0.6742}      & \textcolor{red}{25.46}       & \textcolor{red}{0.6251}       & \textcolor{red}{24.31}        & \textcolor{red}{0.7011}        & \textcolor{red}{26.44}        & \textcolor{red}{0.8211}   \\    \hline
\end{tabular}
}
\label{tab:sr_result}
\end{table*}

Figure \ref{fig:jpeg_combine} gives a qualitative example of the result of our model, where the input image is \textit{woman} from the Set5 \cite{bevilacqua2012low} that is downsampled and compressed by a wide range of quality factors from 10 to 100. It is worth noting that compression with very low quality factors causes a significant color shift to the hue and spatial distribution of the original image, which can be seen in the leftmost LR image (QF = 10). Our model is able to correctly restore the color aberrations of RGB images with a high consistency among different QFs.  

\begin{figure}
    \centering
    \includegraphics[width=1.0\columnwidth]{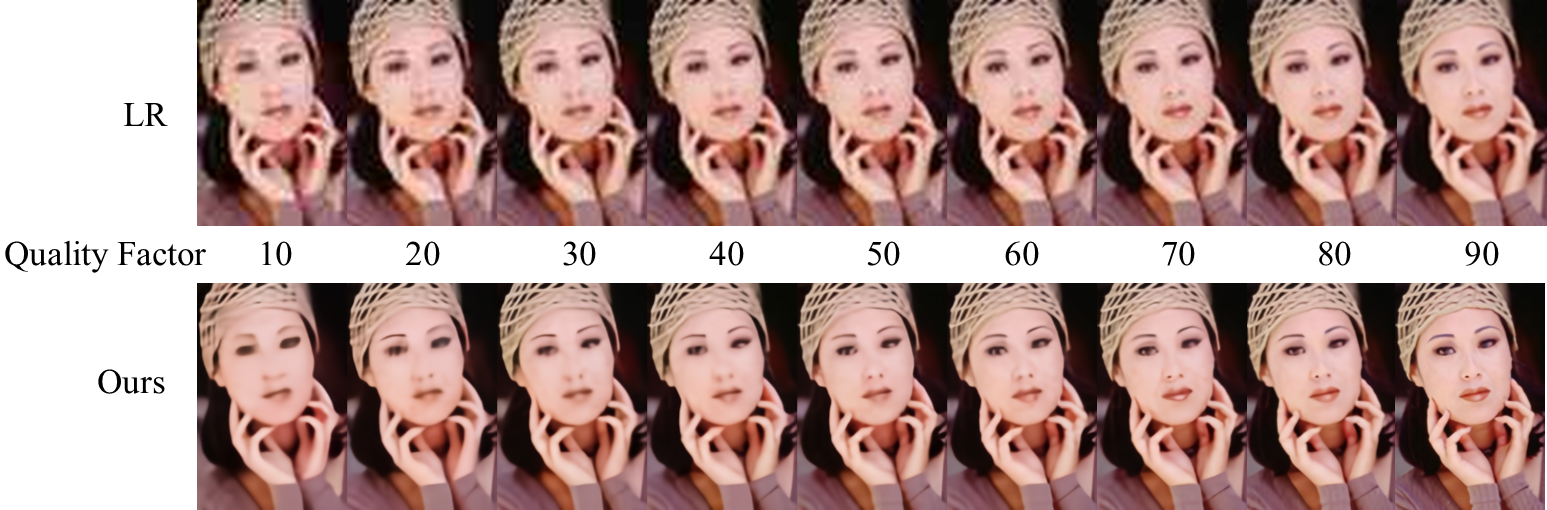}
    \caption{The qualitative result of our network from compressed images with different quality factors (zoom in for a better view). Our model is able to reconstruct reasonable SR images, even at extremely low quality factors. Besides, our results are free of color jittering and other inconsistencies for such a wide range of compression ratios. The image is the ``woman" image from Set5 \cite{bevilacqua2012low}.}
    \label{fig:jpeg_combine}
\end{figure}

\item[Results on User Images]
Besides the above experiments on standard test images, we also conduct experiments on real user images to demonstrate the effectiveness of our model. We mainly focus on the perceptual effect since there are no ground-truth images. Figure \ref{fig:userimage} shows the CAJNN results on real-world image from the WIDER face dataset \cite{yang2016wider}. For comparison, RCAN \cite{zhang2018image} and RRDB \cite{wang2018esrgan} are used as representative SR method, ARCNN \cite{dong2015compression} and DnCNN \cite{svoboda2016compression} are used as the representative CAR methods. The real-world images have unknown downsampling kernels and compression mechanisms, depending on the platforms. According to Figure \ref{fig:userimage}, the SR methods generate images with obvious color shift and ringing artifacts. These artifacts are alleviated with two-stage methods. Still, the results are blurry. Compared with the two-stage methods, our CAJNN can provide SR outputs with sharp edges and rich details, which demonstrates the superiority of our proposed single-stage method when applied to real-world CARSR problems. 


\end{description}

\begin{figure*}[htbp]
	\scriptsize
	\centering
	\begin{tabular}{cc}
        \begin{adjustbox}{valign=t}
			\begin{tabular}{c}
				\includegraphics[width=0.155\textwidth]{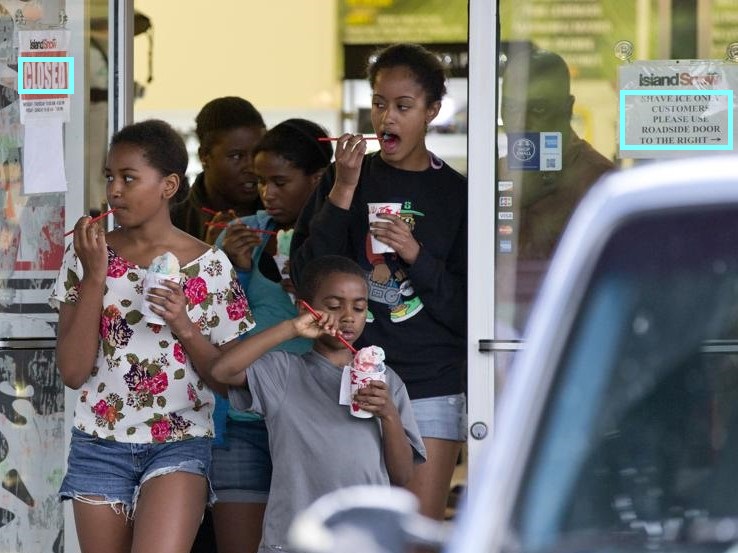}
				\\
				Full-size input
			
			\end{tabular}
		\end{adjustbox}
		\begin{adjustbox}{valign=t}
			\begin{tabular}{ccccccc}
				\includegraphics[width=\widthscale \textwidth]{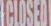} \hspace{-4mm} &
				\includegraphics[width=\widthscale \textwidth]{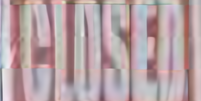} \hspace{-4mm} &
				\includegraphics[width=\widthscale \textwidth]{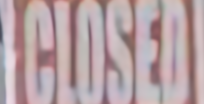} \hspace{-4mm} &
				\includegraphics[width=\widthscale \textwidth]{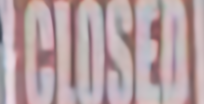}\hspace{-3.5mm} &
				\includegraphics[width=\widthscale \textwidth]{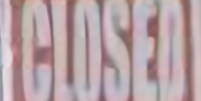}\hspace{-3.5mm} &
				\includegraphics[width=\widthscale \textwidth]{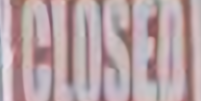}\hspace{-3.5mm} &
				\includegraphics[width=\widthscale \textwidth]{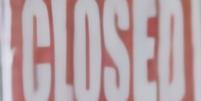}\hspace{-3.5mm}
				\\
				\includegraphics[width=\widthscale \textwidth]{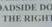} \hspace{-4mm} &
				\includegraphics[width=\widthscale \textwidth]{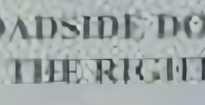} \hspace{-4mm} &
				\includegraphics[width=\widthscale \textwidth]{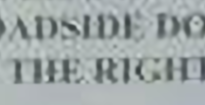} \hspace{-4mm} &
				\includegraphics[width=\widthscale \textwidth]{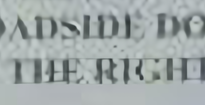}\hspace{-3.5mm} &
				\includegraphics[width=\widthscale \textwidth]{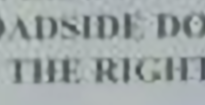}\hspace{-3.5mm} &
				\includegraphics[width=\widthscale \textwidth]{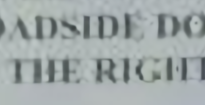}\hspace{-3.5mm} &
				\includegraphics[width=\widthscale \textwidth]{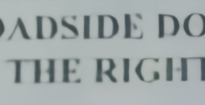}\hspace{-3.5mm}
				\\
				Input \hspace{-4mm} &
    			RCAN \hspace{-4mm} &
				ARCNN+RRDB \hspace{-4mm} &
				ARCNN+RCAN\hspace{-4mm} &
				DnCNN+RRDB\hspace{-4mm} &
				DnCNN+RCAN\hspace{-4mm} &
				\textbf{Ours (CAJNN)}\hspace{-4mm} 
				\\
			\end{tabular}
			\end{adjustbox}
	\end{tabular}
	\caption{CAR \& SR performance comparison of different methods on a user's image from the WIDER face dataset \cite{yang2016wider}. Compared with previous methods, our model can generate artifact-free high-resolution images with sharp edges.}
	\label{fig:userimage}
	\vspace{-5mm}
\end{figure*}

\subsection{Results for Low-Resolution Text Recognition}
Comparing the input LR image and our output in Figure \ref{fig:userimage}, the texts become more readable after being processed by our model. Inspired by this observation, we conducted the following experiments to explore our model's potential to leverage a real-scene text recognition task from low-resolution characters.

We compare the total accuracy of generic text detection on the ICDAR2013 Focused Scene Text dataset \cite{karatzas2013icdar} with TPS-ResNet-BiLSTM-Attn \cite{baek2019wrong} as the text recognition method. The baseline result is acquired by directly recognizing the original input images. As a comparison with the baseline, we use the CAJNN model as described in previous sections to generate artifact-free SR images from the original images and conduct recognition on the output images.

\begin{figure}[htbp]
\captionsetup[subfigure]{labelformat=empty}
\begin{center}
  \begin{subfigure}[b]{0.23\linewidth}
     \includegraphics[width=\linewidth]{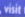}
  \end{subfigure}
  \begin{subfigure}[b]{0.23\linewidth}
  \includegraphics[width=\linewidth]{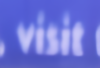}
  \end{subfigure}
  \begin{subfigure}[b]{0.3\linewidth}
  \includegraphics[width=0.7667\linewidth]{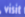}
  \end{subfigure}
  
  \begin{subfigure}[b]{0.23\linewidth}
     \includegraphics[width=\linewidth]{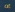}
     \subcaption{GT}
  \end{subfigure}
  \begin{subfigure}[b]{0.23\linewidth}
  \includegraphics[width=\linewidth]{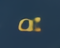}
  \subcaption{Ours ($\times 4$)}
  \end{subfigure}
  \begin{subfigure}[b]{0.3\linewidth}
  \includegraphics[width=0.7667\linewidth]{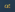}
  \subcaption{Ours (downsampled)}
  \end{subfigure}
\end{center}
\vspace{-5mm}
\caption{Test samples of ICDAR2013 dataset \cite{karatzas2013icdar} (\textit{word\_161}, \textit{word\_836}). The first column shows original input images, the second column is the CARSR output generated by our method, and the third column is acquired by downsampling the second column. By comparing the detection results in the first and second columns, our method can serve as a supportive method for the recognition of low-resolution texts. Besides, the artifact-free image in the third column can also provide more recognizable features for the baseline model without increasing the image size.}
 \label{fig:ocr}
\vspace{-4mm}
\end{figure}

As can be seen in Table \ref{tab:ocr}, the preprocessing of CAJNN improves the recognition accuracy from 85.30\% to 85.75\%, which indicates that the outputs of our model are not only visually appealing to human viewers, but also include more distinct information for the text recognition network as shown in Figure \ref{fig:ocr}. It is worth noting that our output image is $4\times$ the size compared with the baseline inputs, and the average detection time is increased from 31.22s  to 41.56s. Although the improvement in accuracy demonstrates the positive effect yielded by our model, the rise in computation is hard to ignore. Therefore, we disentangle the influence of SR and CAR by bicubicly downsampling the CARSR output images and acquire the third recognition result. Since the image size remains the same as that of the original image, the detection time is identical to the baseline. Compared with the baseline, the recognition accuracy still improves 0.27\% due to the reduction of compression artifacts, which indicates that our model is capable of extracting and maintaining critical features of input images. This experimental result points out a plausible direction for future text recognition research: the image quality plays a vital role in the recognition accuracy, which can be improved by utilizing the learned priors from a pretrained CARSR model.

\begin{table}[htbp]
\caption{Text recognition accuracy on the ICDAR 2013 Focused Scene Text dataset \cite{karatzas2013icdar}. Compared with the baseline method, the introduction of our CARSR method improves the detection performance by 0.45\% (without downsampling) and 0.27\% (with downsampling). }
\vspace{2mm}
\resizebox{\columnwidth}{!}{%
\begin{tabular}{l c c }
\hline
Method & Accuracy & Detection Time (s)\\
\hline 
Baseline \cite{baek2019wrong} & 85.30\%  & 31.22 \\
Ours + Baseline \cite{baek2019wrong} & \textbf{85.75}\%  & 41.56 \\ 
Ours + Downsample + Baseline \cite{baek2019wrong} & 85.57\% & 31.22 \\
\hline
\end{tabular}
}
\label{tab:ocr}
\end{table}

\subsection{Results for Extremely Tiny Face Detection}

Extremely tiny face detection is another practical, yet challenging task in high-level computer vision. Most of the state-of-the-art (SOTA) face detectors \cite{zhu2018seeing,yoo2019extd} for in-the-wild images have already taken various scales and distortions into consideration to achieve impressive detection performance. \cite{bai2018finding} proposed a solution to tackle tiny face detection by explicitly restoring an HR face from a small blurry one using a Generative Adversarial Network (GAN) \cite{goodfellow2014generative}. 

We experimentally validate the effect achieved by our CAJNN on tiny face images in the WIDER FACE dataset \cite{yang2016wider} by comparing the detection results from the following three types of data: original HR (serves as the baseline), downsampled LR (serves as the extremely tiny face inputs), and CARSR outputs from our model. \cite{hu2017finding} is applied as the backbone face detector (We use an unofficial PyTorch \cite{paszke2019pytorch} implementation provided by \hyperlink{https://github.com/varunagrawal/tiny-faces-pytorch}{https://github.com/varunagrawal/tiny-faces-pytorch}). 

\begin{figure*}[htbp]
\captionsetup[subfigure]{labelformat=empty}
\begin{center}
  \begin{subfigure}[b]{0.3\linewidth}
     \includegraphics[width=\linewidth]{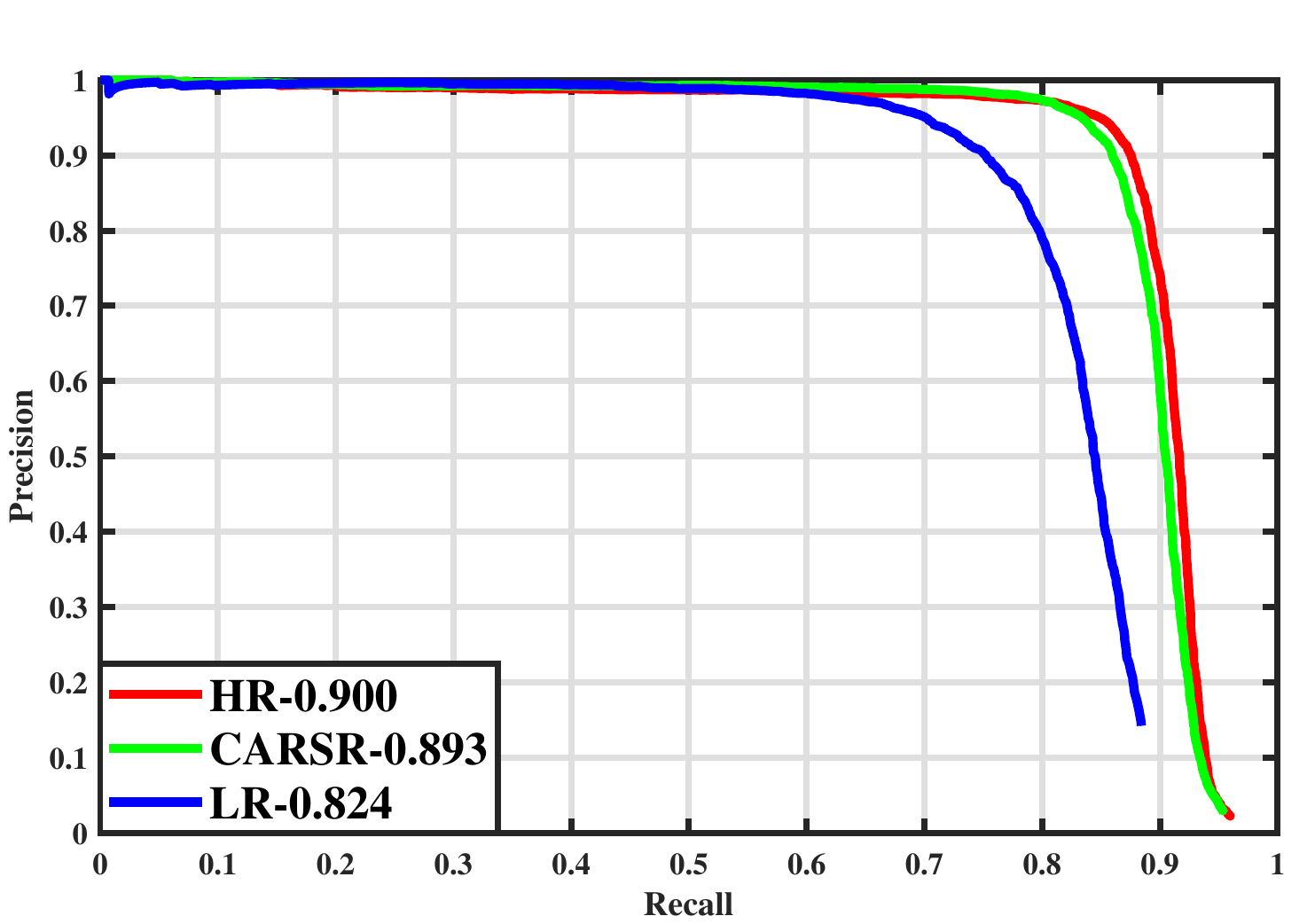}
     \subcaption{Easy}
  \end{subfigure}
  \begin{subfigure}[b]{0.3\linewidth}
  \includegraphics[width=\linewidth]{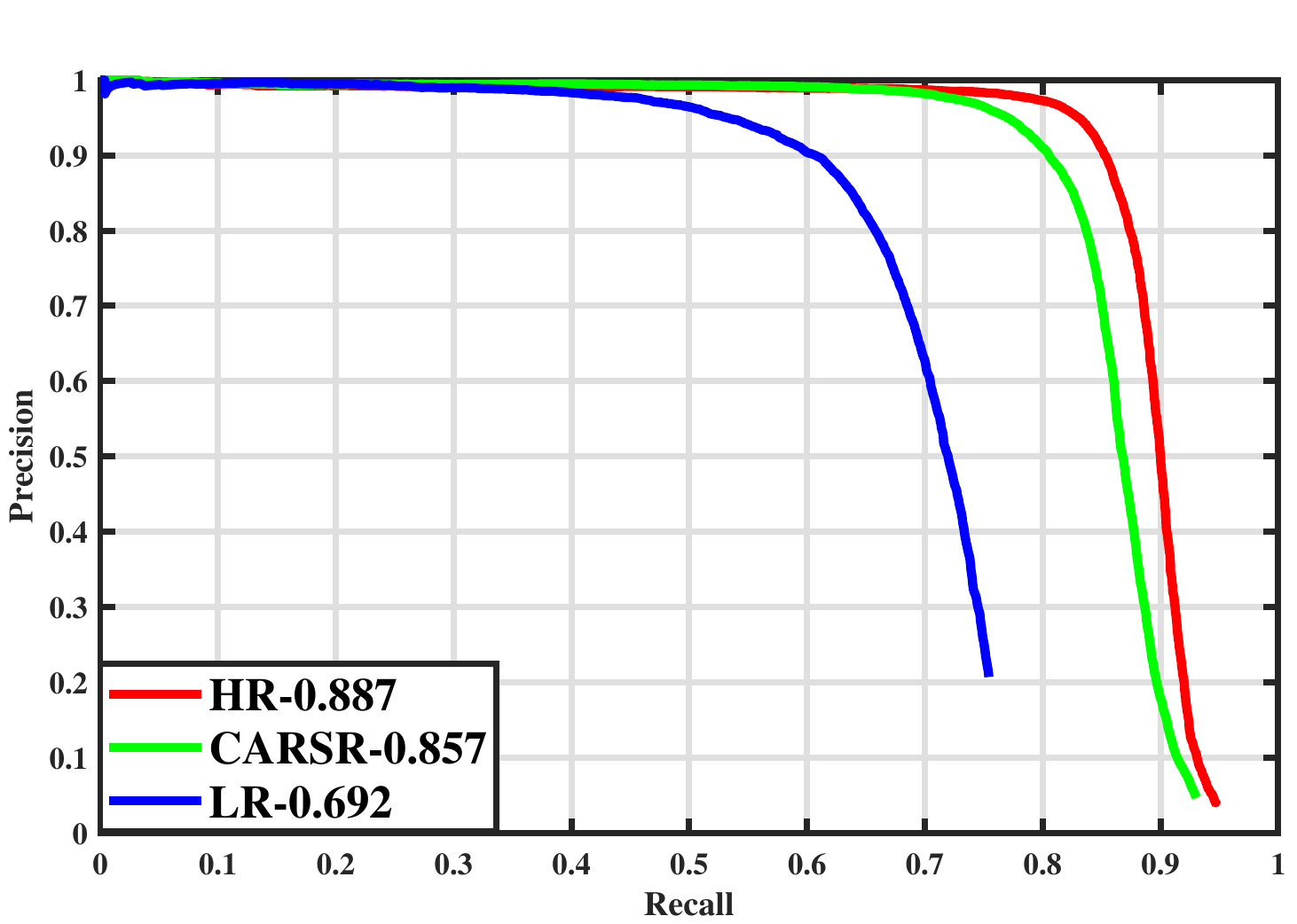}
  \subcaption{Medium}
  \end{subfigure}
  \begin{subfigure}[b]{0.3\linewidth}
  \includegraphics[width=\linewidth]{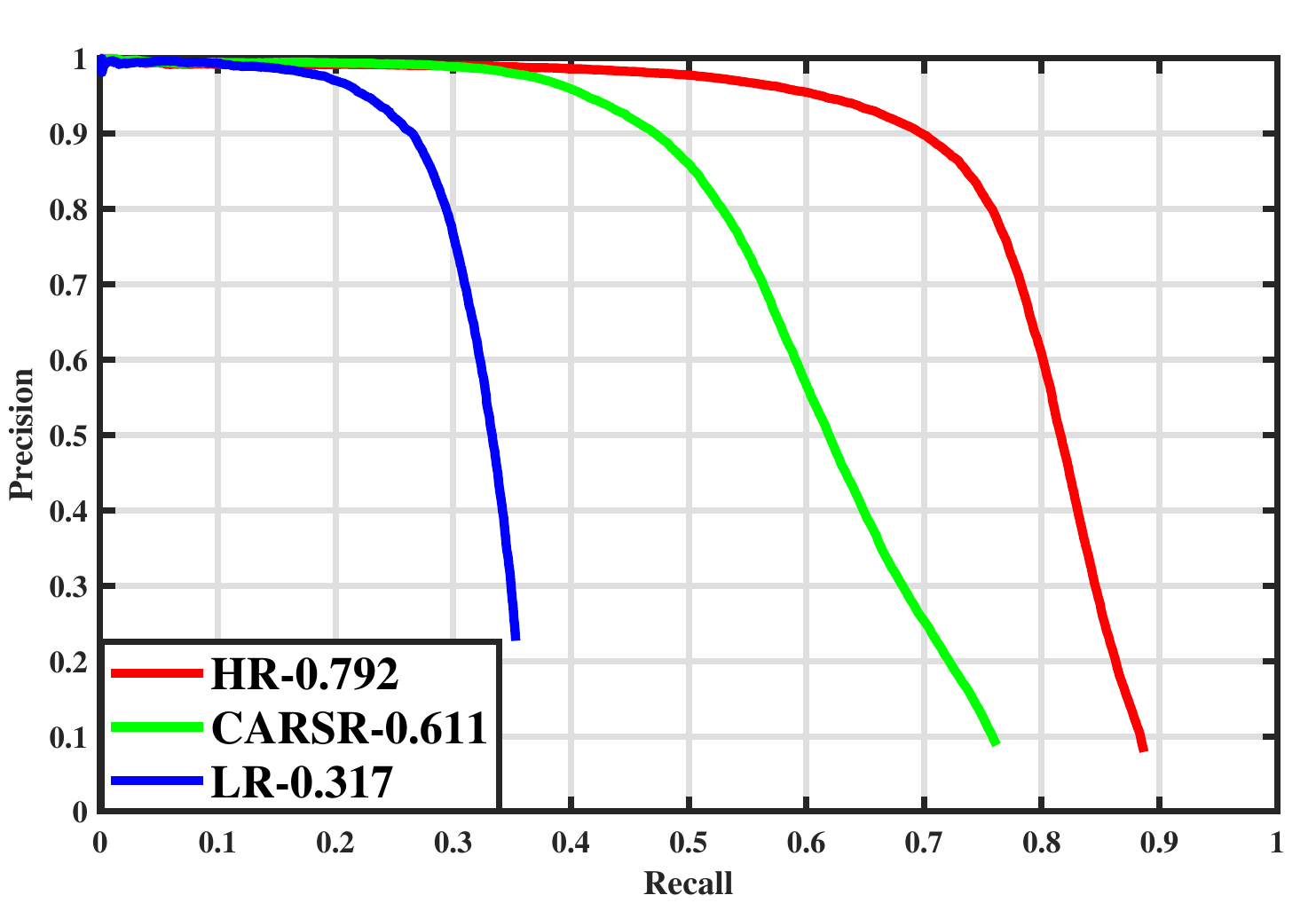}
  \subcaption{Hard}
  \end{subfigure}
\end{center}
\vspace{-5mm}
\caption{The precision-recall curve of three subsets in WIDER FACE \cite{yang2016wider}. The AOC (area under curve) reflects the detector's performance on each type of data (\textcolor{red}{GT}, \textcolor{blue}{LR} and \textcolor{green}{CARSR}). With preprocessing by our model, the detection performance of tiny images can be improved close to that achieved with GT. (Zoom in for a better view.)}
\label{fig:tinyface}
\end{figure*}

Table \ref{tab:tinyface} shows the Average Precision (AP) of the downsampled tiny images and our enhanced ones on all the three validation sets (easy, medium, and hard) of WIDER FACE \cite{yang2016wider}. From Table \ref{tab:tinyface}, we observe that the data processed by CAJNN dramatically improves the detection of LR inputs from 0.317 to 0.611 in AP on the hard set. The reason is that the baseline detector performs downsampling operations by large strides on the tiny faces. Considering the fact that the tiny faces themselves contain less information than average, the detailed information of face structure is lost after several downsampling convolutions. In contrast, our CAJNN provides an artifact-free SR image, which can boost the detection performance by better utilizing the information of small faces. In Figure \ref{fig:tinyface}, the precision-recall curve of our reconstructed image (green line) is close to the ground truth (red line) on the easy and medium subsets. In the hard subset, our CAJNN yields a significant improvement compared to the LR curve. The gap between our output and the GT is due to the irreversible loss of information in extremely tiny faces that happens more frequently in the hard set during the downsampling process.

\begin{table}[tb]
\caption{Average precision of three data types in the WIDER FACE validation set \cite{yang2016wider} with the same face detector \cite{hu2017finding}. The application of our CARSR method greatly improves the detection performance with LR images on all three subsets.}
\begin{center}
\resizebox{0.7\columnwidth}{!}{%
\begin{tabular}{l c c c}
\hline
Input Data & Easy & Medium & Hard\\
\hline 
GT & 0.900 & 0.887 &  0.792 \\
LR & 0.824 & 0.692 & 0.317\\
LR + Ours & 0.893 & 0.857 & 0.611 \\
\hline
\end{tabular}    
}
\end{center}
\label{tab:tinyface}
\end{table}

\subsection{Ablation Study}

\begin{description}[style=unboxed,leftmargin=0cm] 
\item[Effect of Multi-scale Information] As discussed in previous sections, both intra- and inter-block context information is important for designing a CARSR network. In other low-level vision tasks, context information at different scales has already been proved to be effective in improving network performance. Inspired by the first convolution layer of the ResNet \cite{he2016deep}, previous researchers \cite{niklaus2018context} applied $7\times 7$ convolution to extract the context features for the video frame interpolation task. However, such big kernels bring a tremendous number of parameters to the network, especially when embedded in the feature domain, resulting in higher computational cost. Another way of enlarging the filter's receptive field is to use a non-local module \cite{liu2018non,wang2018non}, where the input images are downsampled by convolutional strides and processed at different scales. The non-local module has a rather complex structure and also a large number of parameters. In order to use the context information in a much simpler and lighter representation, our method adopts atrous convolution. By adjusting the dilation rate $r$, the filter can incorporate the context information from a larger receptive field without dramatically increasing the number of parameter as compared to the above methods. 
\end{description}

\begin{table}[tb]
\caption{Ablation Study on the validation set (Set5). We report the performance of CAJNN without the long-range skip connection and ASPP as the baseline. Rows 1-3 show the influence of different ways to extract contextual information by replacing ASPP with other network structures. Rows 4-5 compare the effect of two different upsampling methods on PSNR. The combination of the ASPP and Pixelshuffle modules yields the best performance, and thus is adopted in our network architecture.}
\resizebox{\columnwidth}{!}{%
\begin{tabular}{l c c c c c}
\hline
Model & Base & 1 & 2 & 3 & 4\\
\hline 
Non-local module &  & $\surd$  &  &  &  \\
ASPP &  &  & $\surd$ & $\surd$ & \\
Seqeuntial atrous pooling & &  &  &  & $\surd$ \\ \hline
Upconvolution & $\surd$ & $\surd$ & $\surd$ &  & \\
Pixelshuffle &  &   &  & $\surd$ & $\surd$\\ \hline
PSNR (dB) & 27.868 & 28.274 & 28.276 & \textbf{28.292} & 28.262\\
\hline
\end{tabular}}
\label{tab:sr_ablation}
\end{table}

We conduct an ablation study to illustrate the effect of different ways of representing contextual information in Table \ref{tab:sr_ablation}. In Rows 1-3, we compare the performance of the non-local module, ASPP, and sequential atrous pooling. 
Comparing the base model to Column 1 in Table \ref{tab:sr_ablation}, we can conclude that the introduction of multi-scale information via a non-local module can significantly improve the PSNR by 0.406 dB. This result validates the superiority of aggregating both intra- and inter-block features rather than using a purely local representation for the CARSR task. Furthermore, as seen by comparing Columns 1 and 2, replacing the non-local module by our well-designed ASPP can improve the PSNR by 0.002 dB. Although the improvement is rather small, it is worth noting that the ASPP has fewer convolution layers and parameters, which results in a smaller model size and fewer FLOPs. Remarkably, it can achieve results that are comparable, or even better than that yielded by models with more parameters. By comparing Columns 3 and 4, we also note that the PSNR of ASPP is higher than that of sequential atrous pooling by 0.03 dB, which means that the pyramid-fusion structure is more efficient in representing the multi-scale information. Finally, by comparing Columns 2 and 3 of Table \ref{tab:sr_ablation}, we can observe that the PixelShuffle layer brings a 0.16 dB improvement to PSNR.

\begin{description}[style=unboxed,leftmargin=0cm] 
\item[End-to-End Supervision by Joint CAR and SR] Another ablation study on supervising the CARSR task is conducted to illustrate the effect of joint end-to-end training. Instead of supervising with $I^{HRHQ}$, we attempt to disentangle the CAR and SR by introducing a reconstruction loss according to the definition in Equation \ref{lrhq}, where we can generate an artifact-free LR image $I^{LRHQ}$ from the ground truth $I^{HRHQ}$:
\end{description}
\begin{equation} \label{lrhq}
    I^{LRHQ} = (k \otimes I^{HRHQ})\downarrow_s,
\end{equation}
%

and use it to explicitly supervise the intermediate CAR output $\hat{G}(f^{L'})$ after the context-aware module:
\begin{equation}
     l^{LR} = l(I^{LQHQ}, \hat{G}(f^{L'})).
\end{equation}

Denoting the pixel-wise loss of the final output and ground truth (shown in Equation~\ref{eq:target}) as $l_{HR}$, the overall training loss becomes:
\begin{equation}
    l = l^{HR} + \lambda l^{LR},
\end{equation}
by increasing the weight $\lambda$, we can acquire models trained with higher disentanglement levels. We train three models with $\lambda=0,1,16$ while keeping all the other factors the same. The performance of these models on our validation set is shown in Table \ref{tab:sr_e2e}. The trend is obvious: the PSNR increases as the entanglement increases, which demonstrates the effectiveness of the joint CARSR method with a single-stage network.

\begin{table}
\caption{Ablation Study on joint end-to-end supervision. We introduce the explicit reconstruction loss as a disentanglement mechanism of CAR and SR. By changing the weight of this loss term, we can study the effect of different levels of joint-supervision. Among all the settings, the model trained without the reconstruction loss performs best on our validation set.}
\resizebox{\columnwidth}{!}{%
\begin{tabular}{l c c c}
\hline
Model &  a & b & c \\
\hline 
Weight of reconstruction loss $\lambda$ & 16  & 1 & 0 \\
\hline 
PSNR (dB) & 27.507 & 27.627 & \textbf{27.672} \\ 
\hline
\end{tabular}
}
\label{tab:sr_e2e}
\end{table}

\section{Conclusion}
In this paper, we propose a single-stage network for the joint CARSR task to directly reconstruct an artifact-free high-resolution image from a compressed low-resolution input. To address the CARSR problem, we make use of the contextual information by introducing a specially designed ASPP that integrates both intra- and inter-block features. Our experiments illustrate the effectiveness and efficiency of our method with both standard test images and real-world images. Moreover, the extensive experimental results reveal a high potential for enhancing the performance of current methods for various high-level computer vision tasks, \eg { }real-scene resolution text recognition, and extremely tiny face detection. 



\section*{Acknowledgment}
This research is supported by HP Inc., Palo Alto, CA.

{\small
\bibliographystyle{IEEEtran}
\bibliography{IEEEabrv,egbib}
}

\end{document}